\documentclass[letterpaper,journal]{IEEEtran}
\usepackage{amsmath,amsfonts}
\usepackage{tabularx,booktabs}
\usepackage{algorithmic}
\usepackage{multirow} 
\usepackage{algorithm}
\usepackage{array}
\usepackage[caption=false,font=normalsize,labelfont=sf,textfont=sf]{subfig}
\usepackage{textcomp}
\usepackage{stfloats}
\usepackage{float}
\usepackage{color,xcolor}
\usepackage{url}
\usepackage{verbatim}
\usepackage{graphicx}
\usepackage{bm}
\usepackage{overpic}
\DeclareGraphicsExtensions{.eps}
\usepackage{siunitx}
\usepackage{cuted}
\usepackage{hyperref}

\begin{document}

\title{ADASR:An Adversarial Auto-Augmentation \\ Framework for Hyperspectral and Multispectral Data Fusion}

\author{Jinghui Qin\textsuperscript{$\dagger$}, Lihuang Fang\textsuperscript{$\dagger$}, Ruitao Lu, Liang Lin, and Yukai Shi\textsuperscript{$\ast$}

\thanks{J. Qin, L. Fang, and Y. Shi are with the School of Information Engineering, Guangdong University of Technology, Guangzhou, 510006, China (email: qinjinghui@gdut.edu.cn; 3120002329@mail2.gdut.edu.cn; ykshi@gdut.edu.cn)}

\thanks{R. Lu is with the Department of Missile Engineering, Rocket Force University of Engineering, Xi’an 710025, China (email: lrt19880220@163.com)}

\thanks{L. Lin is with the School of Computer Science and Engineering, Sun Yat-sen University, Guangzhou, 510006, China (email: linliang@ieee.org)
}

\thanks{This work was supported in part by the National Key R\&D Program of China under Grant No.2021ZD0111600, the National Natural Science Foundation of China (NSFC) under Grant No. 62002069 and No. 62206314, Guangdong Basic and Applied Basic Research Foundation under Grant No. 2022A1515011835.}

\thanks{ $\dagger$ The first two authors share equal contributions.}
\thanks{ $\ast$ Corresponding author: Yukai Shi}

}

\maketitle
\begin{abstract}
Deep learning-based hyperspectral image (HSI) super-resolution, which aims to generate high spatial resolution HSI (HR-HSI) by fusing hyperspectral image (HSI) and multispectral image (MSI) with deep neural networks (DNNs), has attracted lots of attention. However, neural networks require large amounts of training data, hindering their application in real-world scenarios. In this letter, we propose a novel adversarial automatic data augmentation framework ADASR that automatically optimizes and augments HSI-MSI sample pairs to enrich data diversity for HSI-MSI fusion. Our framework is sample-aware and optimizes an augmentor network and two downsampling networks jointly by adversarial learning so that we can learn more robust downsampling networks for training the upsampling network.  Extensive experiments on two public classical hyperspectral datasets demonstrate the effectiveness of our ADASR compared to the state-of-the-art methods.
\end{abstract}


\section{Introduction}
\IEEEPARstart{R}{ecently}, there has been a growing interest in developing deep neural networks~\cite{ref8,ref39,ref9,huang2022deep,yang2023fuzzy,jiaxinli} for hyperspectral image (HSI) super-resolution which is a task of producing HSIs from contiguous spectral information in narrow spectral bands. The HSI can be expressed as 3D tensors with 2 spatial dimensions and 1 spectral dimension~\cite{hong2021interpretable}.
Training a neural network robustly often relies on massive and diverse data. However, unlike other image super-resolution tasks with much more synthetic or real training samples, hyperspectral image data is scarce, and the spectral dimension of hyperspectral image data is very high. Therefore, it is non-trivial to train a stable and effective deep neural network. 

Nowadays, data augmentation (DA) is an efficient strategy to lift up the model generalization performance by artificially increasing the volume and diversity of the training data. 
Conventional DA strategies, such as image rotation~\cite{bucci2020effectiveness}, image flip~\cite{shorten2019survey}, etc., often rotate the input image randomly in a pre-defined augmentation angle. Despite its effectiveness on image classification and image super-resolution tasks, this conventional DA approach may lead to insufficient training due to the following reasons: 1) the network training and DA are regarded as two independent phases without joint optimization; 2) the same fixed image rotation augmentation is applied to all input samples without considering the complexity of the samples. Different samples need different rotation angles. Hence, it is insufficient to apply conventional DA to augment the training samples~\cite{graham2014fractional}.

To improve the effect of hyperspectral and multispectral image fusion by augmenting the input samples and training a more stable network, we propose a novel adversarial automatic augmentation framework that jointly optimizes an augmentor network and two downsampling networks, such that the augmentor can learn to produce augmented samples by rotating them at appropriate angles driven by their content to make the two downsampling networks more stable for training upsample network at the next stage. 
Specifically, in the first stage, the augmentor network learns data variations to enrich the input samples by using the loss from a spatial downsampling network and a spectral downsampling network as the feedback. Meanwhile, these downsampling networks take charge of learning a degradation procession on ensuring the generated augmented samples to be semantic-consistent with low-resolution multispectral images so that these downsampling networks can generate appropriate and valuable feedback to optimize the augmentor network. The augmentor network and the downsampling networks are trained in the adversarial learning setting. In the second stage, we train a spectral upsampling network by using the low-spatial-resolution multispectral images generated by the spatial downsampling network and the high spatial resolution multispectral images with reconstruction loss and consistency loss so that we can take full advantage of the priors learned by downsampling networks. The experimental results on two public classical hyperspectral benchamrks demonstrate the effectiveness of our method compared to the state-of-the-art HSI-MSI fusion methods.

\begin{figure*}[t]
  \centering
  \includegraphics[width=0.8\textwidth]{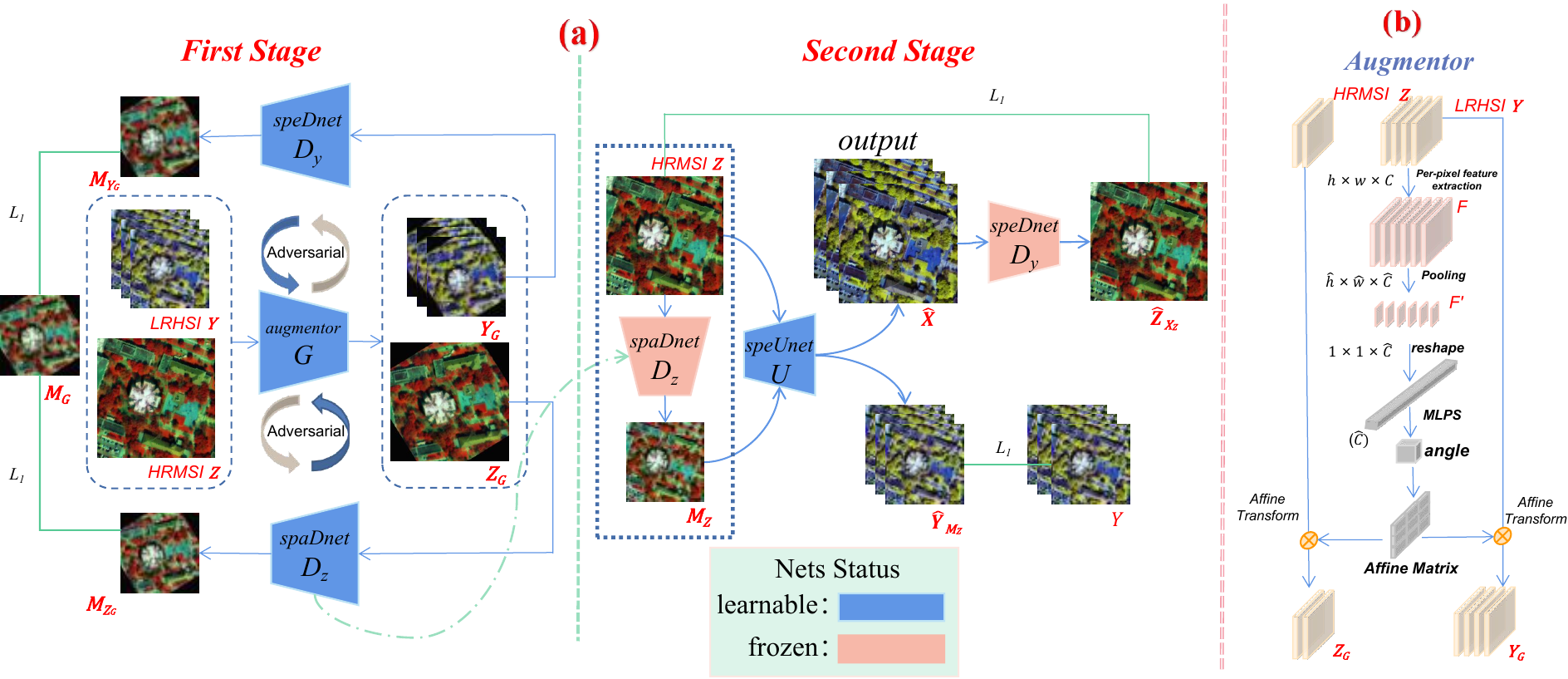}
  \caption{Overview.
  (a) The adversarial auto-augmentation framework: We jointly optimize a data augmentor $G$ and two downsampling networks, a speDnet $D_y$ and a spaDnet $D_z$ at the first stage. 
  The augmentor $G$ learns to generate a sample-specific augmentation function that takes into account the transformation of the image rotation angle, increasing the data diversity for training a better downsampling network. 
  In the second stage, we optimize the spectral upsampling network (SpeUnet) $U$ with the help of the low-resolution multispectral images generated by the fixed downsampling network $D_z$. (b) The design of our data augmentor $G$. The data augmentor $G$ learns to predict the angle to augment the input sample for the degradation model learning. 
  }
  \label{fig:An overview of our image fusion methods and augmentor architecture.}
\end{figure*}

\section{Methodology}
The main contribution of this work is the adversarial auto-augmentation framework that automatically optimizes the augmentation of the input samples for more effective training of the spatial downsampling network. As shown in Fig.~\ref{fig:An overview of our image fusion methods and augmentor architecture.}, our adversarial auto-augmentation framework consists of four modules, including an adversarial data augmentor $G$, a spatial downsampling network (SpaDnet) $D_y$, a spectral downsampling network (SpeDnet) $D_z$, and a spectral upsampling network (SpeUnet) $U$. These modules will be optimized in two training stages. At the first stage, $G$, $D_y$, and $D_z$ are optimized jointly in the adversarial training setting. The augmentor $G$ takes charge of learning a sample-specific rotation angle transformation to increase the data diversity for optimizing downsampling networks $D_y$ and $D_z$ so that we can generate high-quality low-resolution multispectral images for upsampling. Meanwhile, the losses generated by the two downsampling networks will be as feedback to optimize the augmentor $G$. In the second stage, we optimize the speUnet $U$ with the help of the low-resolution multispectral images generated by the fixed spatial downsampling network $D_z$ and the high spatial resolution multispectral images by reconstruction loss and consistency loss.

\subsection{Downsampling Networks}
Let the low-spatial-resolution HSI $\textbf{Y}\in\mathbb{R}^{wh \times C}$ and the low-spectral-resolution MSI $\textbf{Z}\in\mathbb{R}^{WH \times C_{m}}$ be the spatially degraded version and spectrally degraded version of ground-truth HR-HSI $\textbf{X} \in\mathbb{R}^{WH \times C}$. Here, $W$, $H$, and $C$ are the width, height and spectral bands of $\textbf{X}$ while $w$ and $h$ are the width and height of $\textbf{Y}$, and $C_m$ is the spectral band of $\textbf{Z}$ ( $w \ll W$,$h \ll H$, $C_m \ll C$ ), respectively. Then, the degradation models HSI and MSI can be modeled as follows:
\begin{align}
    \textbf{Y} &= \textbf{P}\textbf{X} \label{equ1}, \\
    \textbf{Z} &= \textbf{X}\textbf{S} \label{equ2},
\end{align}
where $\textbf{P} \in\mathbb{R}^{wh \times WH}$ denote the spatial degradation pipeline, which consists of a convolution operation using Point Spread Function (PSF) and a spatial downsample operation, and $\textbf{S} \in\mathbb{R}^{C \times C_m}$ is a band-level spectral response in MSI $\textbf{Z}$. With the $\textbf{Y}$ and $\textbf{Z}$, the HSI-MSI fusion task targets at reconstructing the latent $\textbf{X}$.
In addition, the HSI $\textbf{Y}$'s spectrally degraded result $\textbf{M}_{\textbf{Y}}$ should be equal to the MSI $\textbf{Z}$'s spatially degraded result $\textbf{M}_{\textbf{Z}}$:
\begin{align}
\textbf{M}_{\textbf{Y}} = \textbf{Y}\textbf{S} = \textbf{P}\textbf{Z} = \textbf{M}_{\textbf{Z}}, \label{equ3} 
\end{align}
where $\textbf{M}_{\textbf{Y}} \in \mathbb{R}^{w h \times C m} $ and $\textbf{M}_{\textbf{Z}} \in \mathbb{R}^{w h \times C m}$.

To model the degradation process, we follow prior work~\cite{ref9} to design the spectral downsampling network (SpeDnet) $D_{y}$ with one convolutional layer with the kernels shape $N_k \times C_{i, in} \times C_{i, o} \times 1 \times 1$ and stride size 1 to model the integral procedure with Spectral Response Function (SRF), where $N_k$ denotes both the number of convolutional kernels in SpeDnet and the band number of MSI $\textbf{Z}$. $i$ is the index of the kernels. $C_{i, in}$ is determined by the number of hyperspectral bands that are covered by each band's spectral response in MSI $\textbf{Z}$. $C_{i,o}$ is constrained to 1. That is, each kernel generates only one feature map. $1 \times 1$ is each kernel's spatial size. Therefore, the SpeDnet can be modeled as follows:
\begin{align}
\textbf{M}_{\textbf{Y}(i, j)} = \operatorname{SpeDnet}(\textbf{Y}, \theta)=\frac{\sum_{t \in \Theta_j} \textbf{Y}_{i, t} \omega_j}{\sum \omega_j},
\end{align}
where $\theta$ represents the weights of the $\operatorname{SpeDnet}$, $i$ and $j$ represent the index
of row and column, respectively. $\Theta_j$ represents the $j$-th support set that the band of $\textbf{Y}$ appertains, and $\omega_j$ represents the weights of the $j$-th $C_{in} \times 1 \times 1$ convolution kernel.

Similarly, the spatial downsampling network (SpaDnet) $D_{z}$ is designed to act as PSF. Each band in the spatial dimension is convolved with the same convolutional kernel of size $1 \times r \times r$ and stride $r$, where $r$ ($r = W/w=H/h$) is the spatially dimensional scale factor and the size of convolutional kernels. The SpaDnet can be modeled as follows:
\begin{align}
    \textbf{M}_{\textbf{Z}} =\operatorname{SpaDnet}(\textbf{Z}, \beta),
\end{align}
where $\beta$ is the weight of SpaDnet.

\subsection{Sample Augmentor}
To train the downsampling networks more effectively, the augmentor $G$ learns to generate a sample-specific image rotation angle function for augmenting each input sample.
Our augmentor $G$ takes HSI $\textbf{Y}$ and MSI $\textbf{Z}$ as input and output the augmented samples $\textbf{Y}_{G}$ and $\textbf{Z}_{G}$ respectively. 
The overall architecture of our augmentation procedure is shown in Figure \ref{fig:An overview of our image fusion methods and augmentor architecture.} (b). 
First, a pixel-wise feature extraction unit is deployed to extract features $F \in\mathbb {R}^{\hat{w}\hat{h}\times \hat{C}}$, where $\hat{h}$, $\hat{w}$, and $\hat{C}$ denote the height of the feature map, the width of the feature map, and the number of feature channels, respectively.
Then, the adaptive average pooling operation is applied to build a 
one-pixel and multi-channels feature map $F^{'}$.
Furthermore, a multilayer perceptron takes the $F^{'}$ as input to generate a suitable rotated angle. 
Finally, the augmentor $G$ generates augmented samples with the affine transform and the generated rotation angle. 
Meanwhile, we also use the same rotation angle and affine transform on the ground-truth spatially degraded version $\textbf{M}$, which is given in the training set, so that we can train the downsampling networks and augmentor with adversarial learning strategy. 
We label the augmented $\textbf{M}$ as $\textbf{M}_{G}$. 
The augmented samples $\textbf{Y}_{G}$ and $\textbf{Z}_{G}$ generated by our augmentor can satisfy two following requirements to maximize the network learning: (i) $\textbf{Y}_{G}$ and $\textbf{Z}_{G}$ can be more challenging than $\textbf{Y}$ and $\textbf{Z}$ for downsampling networks since they are rotated and deformed; (ii) $\textbf{Y}_{G}$ and $\textbf{Z}_{G}$ do not lose any semantic information in the original $\textbf{Y}$ and $\textbf{Z}$.

\subsection{Adversarial Learning in First Stage}
To maximize the network learning and generate more challenging samples, we train the augmentor $G$ and the two downsampling networks $D_z$ and $D_y$ with adversarial learning strategy~\cite{ref22}. The augmentor $G$ and the two downsampling networks $D_z$ and $D_y$ are trained alternately and iteratively.

To train the augmentor $G$, $D_z$ and $D_y$ are fixed. The hyperspectral HSI $\textbf{Y}$, high-spatial-resolution MSI $\textbf{Z}$ are fed into data augmentor $G$ and generate new augmented images $\textbf{Y}_G$ and $\textbf{Z}_G$. Then the $\textbf{Y}_G$ is fed into the SpeDnet $D_y$ to generate low-spectral version $\textbf{M}_{\textbf{Y}_G}$ while the $\textbf{Z}_G$ is fed into the SpaDnet $D_z$ to generate low-spatial version $\textbf{M}_{\textbf{Z}_G}$. Finally, we constrain $\textbf{M}_{\textbf{Y}_G}$, $\textbf{M}_{\textbf{Z}_G}$, and $\textbf{M}_{G}$ to be consistent by minimizing the following loss:
\begin{align}
    \mathcal{L}_{G} =  \rho\mathcal{L}\left(\textbf{M}_{\textbf{Y}_G}\right) + \rho \mathcal{L}\left(\textbf{M}_{\textbf{Z}_G}\right), \label{L(G)}
\end{align}
where
\begin{align}
        \mathcal{L}\left(\textbf{M}_{\textbf{Y}_G}\right)
        &= log\left( \frac{1}{whC_m}\left\|\textbf{M}_{\textbf{Y}_G} - \textbf{M}_{G}\right\|_1 \right) \label{L(new_M_YA)},\\
        \mathcal{L}\left(\textbf{M}_{\textbf{Z}_G}\right) &= log\left(\frac{1}{whC_m}\left\|\textbf{M}_{\textbf{Z}_G} - \textbf{M}_{G}\right\|_1 \right) \label{L(new_M_ZA)}, 
\end{align}
$\rho$ is an adjustable hyperparameter.

Similarly, to train the $D_z$ and $D_y$, we fix the augmentor $G$. Then, the original sample $\textbf{Y}$ and its augmented sample $\textbf{Y}_G$ are fed into SpeDnet $D_y$ to obtain $\textbf{M}_{\textbf{Y}}$ and $\textbf{M}_{\textbf{Y}_G}$ while the original sample $\textbf{Z}$ and its augmented sample $\textbf{Z}_G$ are fed into SpaDnet $D_z$ to obtain $\textbf{M}_{\textbf{Z}}$ and $\textbf{M}_{\textbf{Z}_G}$. Finally, we adopt L1 loss to optimize these two downsampling networks as follows:
\begin{equation}
\begin{aligned}
    \mathcal{L}_D &= \left\| \textbf{M}_{\textbf{Y}} - \textbf{M} \right\|_1 + \left\| \textbf{M}_{\textbf{Y}_G} - \textbf{M}_{\textbf{G}} \right\|_1 \\
    &+ \left\| \textbf{M}_{\textbf{Z}} - \textbf{M}\right\|_1 + \left\| \textbf{M}_{\textbf{Z}_G}  - \textbf{M}_{\textbf{G}} \right\|_1,
\end{aligned}  
\end{equation}

\subsection{Spectral Upsample Network in Second Stage}
\label{Spectral upsample network training strategy in stage two}
{The low-spatial-resolution version $\textbf{M}_{\textbf{Z}}$ can be produced by applying the spatial degradation pipeline $\textbf{P}$ in Eq (\ref{equ3}) to the MSI $\textbf{Z}$, where $\textbf{M}_{\textbf{Z}} \in \mathbb{R}^{wh \times C_m}$.}  
From Eq (\ref{equ2}) and Eq (\ref{equ3}), $\textbf{M}_{\textbf{Z}}$ and $\textbf{Z}$ are generated by applying the same spectral degradation operation $\textbf{S}$ to $\textbf{Y}$ and $\textbf{X}$, respectively.
The latent HR-HSI $\textbf{X}$ can be reconstructed if we apply the spectral inverse mapping from ow-spatial-resolution version $\textbf{M}_{\textbf{Z}}$ to hyperspectral HSI $\textbf{Y}$, which is learned in the low resolution, to high-spatial-resolution MSI $\textbf{Z}$.
Therefore, we use the $\textbf{M}_{\textbf{Z}}$ to learn the inverse mapping of the spectrum from $\textbf{M}_{\textbf{Z}}$ to $\textbf{Y}$ by a SpeUnet{. It can be modeled} as follows:
\begin{align}
    \bm{\hat{Y}}_{\textbf{M}_{\textbf{Z}}} &= SpeUnet\left(\textbf{M}_{\textbf{Z}}\right) \label{SpeUnet(MY)},
\end{align}
where the SpeUnet contains 1 × 1 convolution kernels. To optimize the $SpeUnet$, we apply L1 loss to learn spectral inverse mapping as follows:
\begin{equation}
\begin{aligned}
   \mathcal{L}_{U_{1}} = \frac{1}{whC}\left\|   \textbf{Y} - \bm{\hat{Y}_{\textbf{M}_{\textbf{Z}}}}  \right\|_1,
\end{aligned}  
\end{equation}

With an appropriate optimization, HR-HSI $\hat{\textbf{X}}$ can be reconstructed coarsely by inputting $\textbf{Z}$ into $SpeUnet$. However, there exist limitations to the performance of the learned low-resolution spectral inverse mapping. For reconstructing fine-grain HR-HSI $\hat{\textbf{X}}$, we also use the paired $\textbf{Z}$ and $\textbf{X}$ as training data to train the $SpeUnet$ while introducing a consistency loss to optimize $SpeUnet$ for improving its reconstruction performance on high resolution. The consistency loss can be modeled as follows:
\begin{equation}
    \begin{aligned}
        \bm{\hat{Z}}_{\textbf{X}_{\textbf{Z}}} = SpeDnet\left(SpeUnet\left(\bm{Z}\right)\right), \label{SpeDnet(SpeUnet(ZXZ))}
    \end{aligned}
\end{equation}
\begin{equation}
\begin{aligned}
   \mathcal{L}_{U_{2}} = \frac{1}{WHC_m}\left\|   \textbf{Z} - \bm{\hat{Z}_{\textbf{X}_{\textbf{Z}}}}  \right\|_1, \label{LU2}
\end{aligned}  
\end{equation}
Finally, $SpeUnet$ is optimized as follows:
\begin{equation}
\begin{aligned}
   \mathcal{L}_{U} = \mathcal{L}_{U_{1}} +  \alpha\mathcal{L}_{U_{2}},  \label{LU}
\end{aligned}  
\end{equation}
where $\alpha$ is an adjustable hyperparameter.

To obtain the final HR-HSI $\bm{\hat{X}}$, we applied the learned SpeUnet to the original MSI $\textbf{Z}$ as follows:
\begin{align}
    \bm{\hat{X}} = SpeUnet\left(\bm{Z}\right) \label{SpeUnet(ZX)}.
\end{align}

\begin{table*}[htbp]
    \renewcommand{\arraystretch}{1.5}
    \Huge
    \centering
    \caption{QUANTITATIVE PERFORMANCE OF VARIOUS METHODS ON CHIKUSEI AND HOUSTON18 DATASETS}
    \label{tab:CHIKUSEI AND Houston18}
    \setlength{\tabcolsep}{0.7mm}{
      \resizebox{0.8\linewidth}{!}{
        \begin{tabular}{cccccc|ccccc||ccccc|ccccc}
            \cline{1-21}\cline{1-21}
            \multicolumn{1}{c|}{\multirow{3}{*}{Methods}} & \multicolumn{10}{c||}{$\times$ 5}  & \multicolumn{10}{c}{$\times$ 8} \\
            \cline{2-21}
            \multicolumn{1}{c|}{} & \multicolumn{5}{c|}{Chikusei} & \multicolumn{5}{c||}{houston18}  & \multicolumn{5}{c|}{Chikusei} & \multicolumn{5}{c}{houston18} \\\cline{2-21}
            \multicolumn{1}{c|}{}        &  SAM$\downarrow$     &  ERGAS
            $\downarrow$  &  PSNR$\uparrow$      &  RMSE$\downarrow$      &   CC$\uparrow$   &  SAM$\downarrow$     &  ERGAS$\downarrow$  &  PSNR$\uparrow$      &  RMSE$\downarrow$      &   CC$\uparrow$
            &  SAM$\downarrow$     &  ERGAS$\downarrow$  &  PSNR$\uparrow$      &  RMSE$\downarrow$      &   CC$\uparrow$   &  SAM$\downarrow$     &  ERGAS$\downarrow$  &  PSNR$\uparrow$      &  RMSE$\downarrow$      &   CC$\uparrow$
            \\\hline\hline
            \multicolumn{1}{c|}{HySure}  & 1.1953 & 1.0697 & 42.4036 & 0.0081 & 0.9974  & 1.5485 & 0.6756 & 43.0993 & 0.0052 & 0.9989
                                         & 1.5365 & 0.8894 & 40.0119 & 0.0107 & 0.9968  & 2.1120 & 0.5719 & 40.2963 & 0.0070 & 0.9984\\
                                         
            \multicolumn{1}{c|}{FUSE}    & 1.3413 & 1.2055 & 41.1566 & 0.0094 & 0.9967  & 1.6768 & 0.9117 & 40.5935 & 0.0072 & 0.9979
                                         & 1.4428 & 0.8323 & 40.5532 & 0.0098 & 0.9972  & 1.8498 & 0.5673 & 40.7906 & 0.0066 & 0.9980 \\
            
            \multicolumn{1}{c|}{G-SOMP+} & 1.2874 & 1.3306 & 41.2452 & 0.0091 & 0.9945  & 1.4909 & 0.6880 & 42.8371 & 0.0054 & 0.9989
                                         & 1.5247 & 1.0934 & 38.8789 & 0.0117 & 0.9914  & 1.8978 & 0.5429 & 40.5534 & 0.0068 & 0.9984\\
                                         
            \multicolumn{1}{c|}{CSU}     & 1.4397 & 1.7031 & 39.8464 & 0.0097 & 0.9898  & 1.4980 & 0.6912 & 42.5571 & 0.0054 & 0.9987
                                         & 1.6817 & 1.1877 & 38.3814 & 0.0117 & 0.9879  & 1.9006 & 0.5476 & 40.3440 & 0.0068 & 0.9981\\
                                         
            \multicolumn{1}{c|}{CNMF}    & 1.0918 & 1.0752 & 42.6555 & 0.0079 & 0.9964  & 1.2197 & 0.6054 & 43.9170 & 0.0047 & 9.9991
                                         & 1.2458 & 0.9126 & 39.8480 & 0.0105 & 0.9951  & 1.5856 & 0.4972 & 41.3301 & 0.0063 & 0.9986\\ 
                                         
            \multicolumn{1}{c|}{STEREO}  & 0.8801 & 0.8282 & 49.7968 & 0.0043 & 0.9968  & 1.0094 & \textcolor{blue}{0.3691} & \textcolor{blue}{51.0273} & \textcolor{blue}{0.0028} & 0.9994
                                         & 1.0282 & 0.5957 & 48.8520 & 0.0050 & 0.9958  & 1.1090 & \textcolor{blue}{0.2525} & \textcolor{blue}{50.4403} & \textcolor{blue}{0.0030} & 0.9992\\
                                         
            \multicolumn{1}{c|}{CSTF}    & 1.2306 & 1.3534 & 45.1193 & 0.0059 & 0.9932  & 1.4285 & 0.5509 & 46.9136 & 0.0037 & 0.9988
                                         & 1.2458 & 0.8431 & 45.1089 & 0.0060 & 0.9930  & 1.4539 & 0.3513 & 46.8232 & 0.0038 & 0.9988\\ 
                                         
            \multicolumn{1}{c|}{DHIF-Net} & 1.4113 & 1.6151 & 39.6355 & 0.0096 & 0.9904 & 1.4108 & 0.6578 & 42.8076 & 0.0052 & 0.9987
                                          & 1.7132 & 1.3255 & 37.7378 & 0.0118 & 0.9841 & 1.7075 & 0.5127 & 40.8392 & 0.0064 & 0.9979 \\ 
                                          
            \multicolumn{1}{c|}{CUCaNet} & 1.0353 & 0.8356  & 48.4793 & 0.0054 & 0.9973 & 1.7031 & 0.7984 & 42.2861 & 0.0066 & 0.9986
                                         & 0.8561 & 0.4843  & 49.6982 & 0.0044 & 0.9974 & 1.7450 & 0.4818 & 43.2826 & 0.0064 & 0.9987\\

            \multicolumn{1}{c|}{UDALN}   & \textcolor{blue}{0.7127} & \textcolor{blue}{0.6851} & \textcolor{blue}{52.3858} & \textcolor{blue}{0.0037} & \textcolor{blue}{0.9980}  
                                         & \textcolor{blue}{0.8770} & 0.6698 & 42.6841 & 0.0053 & \textcolor{blue}{0.9994}
                                         & \textcolor{blue}{0.7504} & \textcolor{blue}{0.4574} & \textcolor{blue}{49.2979} & \textcolor{blue}{0.0045} & \textcolor{blue}{0.9976}
                                         & \textcolor{blue}{0.8896} & 0.5452 & 40.2321 & 0.0070 & \textcolor{blue}{0.9994} \\\hline\hline 

            \multicolumn{1}{c|}{ADASR(OUR)}     
            &\textcolor{red}{0.7032}   & \textcolor{red}{0.5742} & \textcolor{red}{53.6463} &\textcolor{red}{0.0035}  & \textcolor{red}{0.9983} 
            & \textcolor{red}{0.8234}  & \textcolor{red}{0.3132} & \textcolor{red}{53.5342} & \textcolor{red}{0.0024} & \textcolor{red}{0.9995}
            &\textcolor{red}{0.7395}   & \textcolor{red}{0.4347} & \textcolor{red}{52.0179} & \textcolor{red}{0.0038} & \textcolor{red}{0.9977}
            & \textcolor{red}{0.8438}  & \textcolor{red}{0.2007} & \textcolor{red}{53.1627} & \textcolor{red}{0.0025} & \textcolor{red}{0.9995} \\
            \cline{1-21}\cline{1-21}
        \end{tabular}
      }
    }
\end{table*}
\section{Experiments}
 
 \subsection{Datasets, Baselines, and Metrics\label{Datasets and fusion methods}}
\noindent\textbf{Datasets}. 
{To evaluate the efficiency of our design strategy, we adopted two widely used hyperspectral datasets. The first one called Houston18 for short has 48 bands with wavelengths ranging from 380 to 1050 nm. This dataset contains $1202 \times 4172$ pixels with a spatial resolution of 1 m. 
The second one, named Chikusei for short, has 128 bands with wavelengths ranging from 363 to 1018 nm and contains $2517 \times 2335$ pixels with a spatial resolution of 2.5 m. 
After removing some noisy and water absorption bands, the sub-images of $240 \times 240 \times 46$ on Houston18 and $240 \times 240 \times 110$ on Chikusei are chosen for the test.}
The two sub-images are acted as reference images for comparison. In synthesizing the RGB images from the Houston18 and Chikusei datasets, we used 46th and 110th spectral bands as \textcolor{red}{red}-band, 30th and 75th spectral bands as \textcolor{green}{green}-band, and 14th and 30th spectral bands as \textcolor{blue}{blue}-band.

\noindent\textbf{Baselines}. To demonstrate the effectiveness of our ADASR, we compare our ADASR with 10 SOTA HSI-MSI fusion methods, including 1 Bayesian representation based method (FUSE~\cite{ref2}), 4 matrix factorization based methods (HySure~\cite{ref3}, CNMF~\cite{ref4}, CSU~\cite{ref5} and G-SOMP+~\cite{ref6}), 2 tensor factorization based methods (CSTF~\cite{ref7} and STEREO~\cite{ref8}), 1 supervised deep learning method (DHIF-Net~\cite{huang2022deep}), 2 unsupervised deep learning methods (CUCaNet~\cite{ref39} and UDALN~\cite{ref9}). We abbreviate our method as ADASR in the following description.

\noindent\textbf{Metrics}. 
We deploy 5 metrics, including spectral angle mapper (SAM), error relative to the global adimensional synthesis (ERGAS), peak signal-to-noise ratio (PSNR), root mean square error (RMSE), and correlation coefficients (CC) to evaluate the performance.

\subsection{Implementation Details\label{Implementation details}}
The ADASR is implemented by using PyTorch~\cite{ref21}, and trained on the Linux server with an NVIDIA Titan XP GPU.
We set the training step as 40,000 and deployed the Adam~\cite{chen2016bridging} optimizer with a learning rate of 0.0001. In addition, in Equation $\ref{LU}$, we set the parameter $\alpha$ to 0.3. To train the downsampling networks, we take the HSI and MSI pairs as input and use trainable $PSF$ and $SRF$ for downsampling networks. To reduce model oscillations~\cite{ref22}, we follow the prior work~\cite{ref37} and train the downsampling networks $D_z$ and $D_y$ by using mixed original and augmented training samples rather than using original training samples only. 

\begin{figure}[t]
  \centering
  \includegraphics[width=0.8\linewidth]{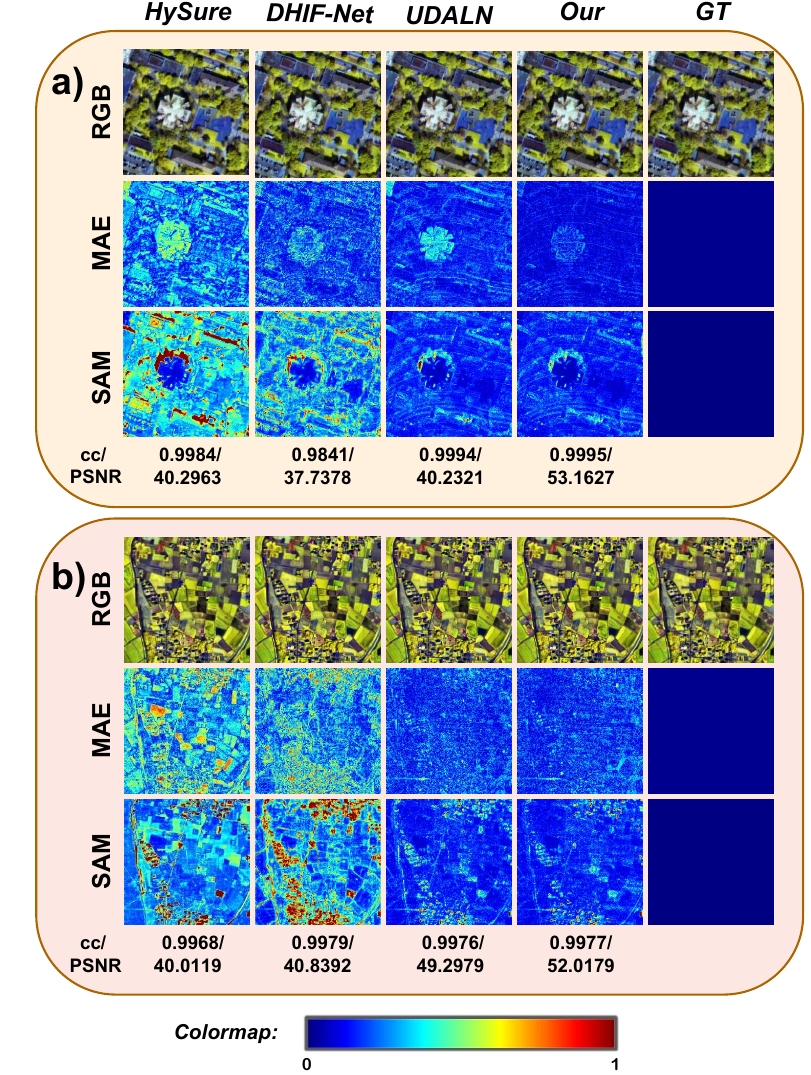}
  \caption{Visual results generated by different methods on the Housoton18 and Chikusei datasets in the scale factor 8. The 1st row shows the reconstruction results for the whole image, the 2nd row shows the MAE heatmap, and the 3rd row shows the SAM error heatmap. 
  }
  \label{fig:visual result}
  \vspace{-5mm}
\end{figure}

\subsection{Main Results\label{main results}}
The quantitative results for the Housoton18 and Chikusei datasets with scale factors of 5 and 8 are shown in Table \ref{tab:CHIKUSEI AND Houston18}. From the results, we can observe that our method outperforms all baselines on both two datasets. The improvements (\%) on SAM/ERGAS/PSNR/RMSE/CC metrics for Houston18 and Chikusei datasets in the scale 8 are $5.1/20.5/5.4/16.7/0.01$ and $1.5/5.0/5.5/15.6/0.01$, respectively. Meanwhile, the improvements (\%) in the scale 5 are $6.1/15.1/4.9/14.3/0.01$ and $1.3/16.2/2.4/5.4/0.03$.
These quantitative improvements demonstrate the superiority of our method. Besides, we also conduct the qualitative analysis by showing the synthetic RGB images of HR-HSI and the mean absolute error (MAE) heatmap and SAM heatmap between the reconstructed HR-HSI and the reference HR-HSI as in the Fig. \ref{fig:visual result}. We can see that our method can achieve better SAM and MAE. The qualitative results also show the reconstruction superiority of our method to reconstruct details more efficiently.

\subsection{Ablation Study \label{Ablation Study}}

\subsubsection{The effects of the data augmentation and the consistency loss}
In our framework, the data augmentor $G$ and the consistency loss $\mathcal{L}_{U_{2}}$ are two key components, so we conduct ablation studies on these two components. The experimental results are shown in Table \ref{tab:Ablation study of our method.}. We can observe that either the auto data augmentation or the consistency loss can improve the performance, while the best performance can be achieved by applying both auto data augmentation and the consistency loss.

\begin{table}[h!]
    \renewcommand{\arraystretch}{1.2}
    \centering
    \footnotesize
    \caption{Ablation study of different components. `-' means the component is removed
    }
    \setlength{\tabcolsep}{2.5mm}{
    \resizebox{0.8\linewidth}{!}{
    \begin{tabular}{c|ccccc}
       \toprule
       \multicolumn{1}{c|}{\multirow{2}{*}{Model}} & \multicolumn{5}{c}{Metric}      \\ \cline{2-6}
       \multicolumn{1}{c|}{}      &  SAM$\downarrow$     &  ERGAS$\downarrow$  &  PSNR$\uparrow$      &  RMSE$\downarrow$      &   CC$\uparrow$    \\\hline\hline
       -$G$-$\mathcal{L}_{U_{2}}$                        & 0.8655      & 0.5891 &  43.9732      & 0.0046    &0.9995     \\
       -$G$                        &  0.8591    &  0.3602  &  52.3446     &  0.0028   &0.9995 \\
       -$\mathcal{L}_{U_{2}}$                          &  \textcolor{blue}{0.8255}   &  \textcolor{blue}{0.3160}  & \textcolor{blue}{53.5126}        &  \textcolor{blue}{0.0024}   &\textcolor{blue}{0.9995}   \\\hline\hline
      ADASR & \textcolor{red}{0.8234}  & \textcolor{red}{0.3132} & \textcolor{red}{53.5342} & \textcolor{red}{0.0024} & \textcolor{red}{0.9995} \\
       \bottomrule
    \end{tabular}}}
    \label{tab:Ablation study of our method.}
\end{table}
\subsubsection{Can the learned augmentor $G$ work better?}
We also explore whether our learned augmentor $G$ can work better than the conventional image augmentation method - random rotation augmentation or without any image augmentation. The results on the Houston18 dataset are shown in Table \ref{tab:Study of the effectiveness of data augmentor}. We can observe that the conventional image augmentation method can not improve the performance, while our method can improve the performance on all metrics. These results show the effectiveness of our adversarial auto-augmentation framework.

\begin{table}[ht]
    \renewcommand{\arraystretch}{1.2}
    \centering
    \scriptsize
    \caption{Study of the effectiveness of data augmentor}
    \setlength{\tabcolsep}{2.5mm}{
    \resizebox{0.8\linewidth}{!}{
    \begin{tabular}{c|ccccc}
        \toprule
        \multirow{2}{*}{Method}& \multicolumn{5}{c}{Metric}      \\ \cline{2-6}
        \multicolumn{1}{c|}{}   &  SAM$\downarrow$     &  ERGAS$\downarrow$  &  PSNR$\uparrow$ &  RMSE$\downarrow$      &   CC$\uparrow$    \\\hline\hline
        No augmentation             & 0.8591 & 0.3602 & 52.3446 & 0.0028 & 0.9995  \\
        Random rotation      & 0.8452 & 0.3351 & 53.0553 & 0.0025 & 0.9994  \\\hline\hline
        ADASR   & \textcolor{red}{0.8234}  & \textcolor{red}{0.3132} & \textcolor{red}{53.5342} & \textcolor{red}{0.0024} & \textcolor{red}{0.9995}
        \\
        \bottomrule
    \end{tabular}}}
    \label{tab:Study of the effectiveness of data augmentor}
\end{table}

\section{Conclusion \label{Conclusion}}
In this letter, to improve the effect of hyperspectral and multispectral image fusion by augmenting the input samples and training a more stable network, we propose a novel adversarial automatic augmentation framework ADASR that jointly optimizes an augmentor network and two downsampling networks so that the augmentor network can augmented samples automatically by rotating them at appropriate angles driven by their content to make the two downsampling networks more stable for training upsample network at the next stage.  
Specifically, the augmentor network and the downsampling networks are trained by reconstructing low-spatial resolution multispectral images in the adversarial learning setting. Then, we train a spectral upsampling network by both high spatial resolution multispectral images and their generated low spatial resolution multispectral images with reconstruction loss and consistency loss, so that we can take full advantage of the priors learned by downsampling networks. The experimental results on two public classical hyperspectral datasets demonstrate the effectiveness of our ADASR compared to the state-of-the-art HSI-MSI fusion methods.

\bibliographystyle{IEEEtran}
\bibliography{reference}

\end{document}